\documentclass[]{article}
\usepackage[letterpaper]{geometry}
\usepackage{amta2020}
\usepackage{times}
\usepackage{url}
\usepackage{latexsym}
\usepackage{natbib}
\usepackage{layout}

\usepackage{graphicx}

\usepackage{booktabs}
\usepackage{algorithm}
\usepackage{algorithmic}

\begin{document}

\title{\bf Machine Translation System Selection\\from Bandit Feedback}

\author{\name{\bf Jason Naradowsky} \hfill  \addr{narad@preferred.jp}\\
        \addr{Preferred Networks, Tokyo, Japan}
\AND
        \name{\bf Xuan Zhang} \hfill \addr{xuanzhang@jhu.edu}\\       \name{\bf Kevin Duh} \hfill \addr{kevinduh@cs.jhu.edu}\\
        \addr{Johns Hopkins University,
        Baltimore, USA}
}

\maketitle
\pagestyle{empty}

\begin{abstract}
Adapting machine translation systems in the real world is a difficult problem.  In contrast to offline training, users cannot provide the type of fine-grained feedback (such as correct translations) typically used for improving the system.  Moreover, different users have different translation needs, and even a single user's needs may change over time.

In this work we take a different approach, treating the problem of adaptation as one of selection.  Instead of adapting a single system, we train many translation systems using different architectures, datasets, and optimization methods.  Using bandit learning techniques on simulated user feedback, we learn a policy to choose which system to use for a particular translation task.  We show that our approach can (1) quickly adapt to address domain changes in translation tasks, (2) outperform the single best system in mixed-domain translation tasks, and (3) make effective instance-specific decisions when using contextual bandit strategies.  
\end{abstract}

\section{Introduction}

Recent advances in machine translation have greatly improved translation quality on in-domain data~\citep{vaswani:2017}. 
But choosing the best system to deploy for a given translation task can be difficult, as many different systems could be considered approximately state-of-the-art, and there is not a single system which is best for all situations. 
For instance, while neural machine translation (NMT) traditionally excels in big data scenarios, when data is scarce it is not uncommon for a statistical phrased-based translation (SMT) to be the better choice.  Even different hyperparameter settings of the same model may yield systems which each excel at different translations tasks.  

In this work we explore the practical question of how best to deploy and improve an MT service over time. One solution to this problem is adaptation, where the model continues to train in an online manner during deployment, 
and the model parameters are updated in response to new types of data.  
However, adapting a system in this manner has the potential to cause catastrophic forgetting~\citep{kirkpatrick2016overcoming}: the model parameters shift too much, and performance on the original translation task declines.

A second practical concern is that for translation services deployed in the real world, the degree of feedback is often limited.  Users of Google Translate can rate the quality of a translation as ``helpful'' or ``wrong'', and Facebook users can use an ordinal scale from 1 to 5, but neither can realistically ask a user to provide a reference translation. 
Furthermore, the user provides feedback only a single time per translation, and multiple users are unlikely to ask for translations of the same text.  To what extent can we leverage such feedback to quickly adapt our system to the user's translation needs?

Here we turn to \emph{bandit learning}, a class of strategies for learning a decision policy in an online setting from such limited feedback.  We assume access to a number of pre-trained machine translation systems, which vary in terms of architecture, training data, and optimization method.  The policy must then determine which translation system to use for a given source sentence.  In contrast to adaptation, the parameters of each translation system are fixed, preventing the possibility of catastrophic forgetting. 

\begin{figure*}
\centering
\includegraphics[scale=0.75]{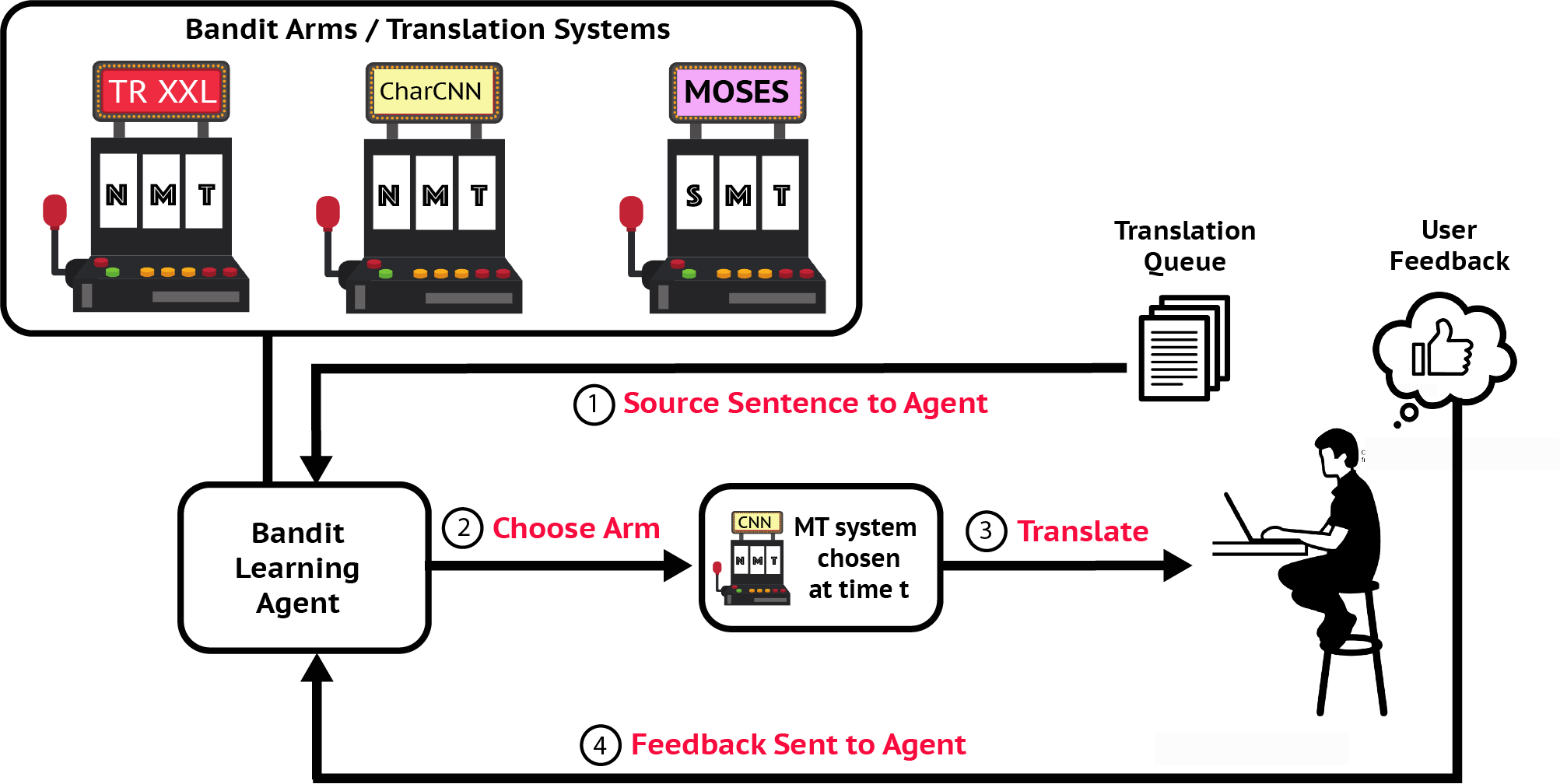}
\caption{\label{fig:bandit} Basic setup: Bandit Learning Agent selects the MT system to use at time $t$. Based on user feedback (e.g. thumbs up/down or other signals), the bandit learning agent adapts to the stream of successive source sentences. The goal is to adapt quickly from limited user feedback.}
\end{figure*}

The bandit learning setup is illustrated in Figure~\ref{fig:bandit}. We assume no prior knowledge of the domain(s) of sentences that the user wants to translate. Source sentences are fed to the system in sequence, and at each time step the bandit learning agent chooses one pre-built MT system to generate translations. The user provides simple feedback for each translation (such as a thumbs up/down), and the goal of the agent is to converge to the best MT system as quickly as possible.

We compare several bandit methods across three data domains (and mixtures of these domains), 
and find that:
\begin{itemize}
    \item Even simple bandit algorithms can quickly adapt to new domains, and converge to choosing optimal/near-optimal systems after a few hundred examples. 
    \item For the case of contextual bandits, where we can condition on a particular source sentence when making a decision, simple features derived from sentence length, vocabulary, and BERT, are effective.  In comparison to an oracle which chooses the single best arm for a given test set, our contextual system can vastly outperform it on mixed-domain settings.
    \item The methods are robust to different forms of simulated human feedback proposed in the machine translation literature.
\end{itemize}
We present bandit-based translation system selection as a viable alternative to deploying or adapting a single MT system.

\section{Bandit Learning}

We now provide a brief overview of bandit algorithms.  Borrowing terminology from casino slot-machines, which are sometimes referred to as ``one-armed bandits'', a bandit problem presents the gambler with a choice: given a bandit with multiple arms, which should be pulled to maximize overall earnings?  Assume that each arm has its own payoff distribution.  Even though this distribution is not observed, the gambler may start preferring a particular arm over time, associating it with higher reward than others.   

Formally, let there be a $K$-arm bandit, where each arm corresponds to an action. At each timestep $t$, an agent must choose an action $k$, corresponding to selecting an arm $1 \leq k \leq K$. Each arm $k$ is associated with a reward $r_k^t$, where higher values are desired.  
The agent's goal is to minimize cumulative regret (the amount of reward lost by making suboptimal decisions) over the course of $T$ timesteps:
\begin{equation}
\label{eq:regret}
   Regret = \sum_t^T (\max_{k' \in K}\ r_{k'}^t) - r_\pi^t
\end{equation}

\noindent where $r_\pi^t$ denotes the reward of the arm chosen by the agent's policy, $\pi$.
Note the agent does not necessarily minimize regret in the sense that it is an objective function to optimize. It does not have access to the oracle action nor the value $(\max_{k'}\ r_{k'}^t)$; instead, the agent receives information only on the arm it chooses ($r_\pi^t$). Intuitively, one can imagine that the agent needs to try different actions in order build up a profile of rewards for each arm.  

Thus a K-arm bandit is a classic exploration vs. exploitation problem.  Exploring new or infrequent actions improves our understanding of their effects, and may lead us to discover better strategies.  However, doing so comes at the cost of not exploiting the actions we currently believe to be the best.  There are many established algorithms for solving bandit problems, each approaching this trade-off in a different way.  We introduce some of these methods later in Section \ref{sec:bandit_methods}, before applying them to the practical problem of simulated human-in-the-loop translation system selection.

\section{Translation Bandits}

We now define online translation system selection as a bandit learning problem.  In our formulation, each arm is a translation system, i.e., pulling a bandit arm is equivalent to selecting a pre-trained translation system and applying it to a given source sentence.  Learning takes place across a series of rounds, and our goal is to continually adapt the choice of translation system to changing user needs (represented by the source domain), thereby maximizing user satisfaction. 

We assume that the source sentences that need to be translated are in a queue, and we process them in sequential order. 
The bandit selection process (at time $t$) is as follows:
\begin{enumerate}
    \item Observe a source sentence $s^t$
    \item (Optionally) Compute features $\phi(s^t)$
    \item Choose a MT system $k^t = \pi(s^t)$ and generate a translation $g^t = k^t(s^t)$ for the user
    \item The user gives feedback in the form of a reward $r_\pi^t$ based on the quality of the translation. This can be, for example, a thumbs up/down rating. In our \textit{simulation} experiments, we compute reward as $e^t = SentenceBLEU(s^t, g^t)$ based on reference translation $g^t$ and perturb it to approximate coarse, noisy human feedback.\footnote{We compute sentence-level BLEU because rewards are defined per example for standard bandits. Nevertheless, our final evaluation is in terms of corpus-level BLEU. Note that we use perturbed SentenceBLEU only as a way to simulate human feedback; we do not assume sentence-level feedback to be consistent with corpus-level metrics.} 
    \item Update selection policy $\pi$
\end{enumerate}

\paragraph{Contextual Bandits}
One of the major distinctions between classes of bandit algorithms is whether the policy can condition its decision on a time-specific observation.  Methods which do are referred to as \emph{contextual bandits}, and are able to make more nuanced, instance-specific decisions.  In other words, whereas simple bandits attempt to learn which arm is best, contextual bandits also learn \emph{when} it is best.  However, this necessitates computing a feature representation from the source sentence (Step 2). 

\paragraph{Simulated Feedback}
In a real world deployment, user satisfaction is obtained directly from the user, but for the purpose of conducting this study in a tractable and repeatable manner, we simulate the human-in-the-loop. 
In this simulated setting we have access to reference translations, and can compute the true SentenceBLEU score (Step 4).  However, using such a metric would not be a realistic approximation of real user feedback.  Previous work \citep{nguyen-daumeiii-boydgraber:2017:EMNLP2017} identified several ways in which human judgements may differ from more traditional continuous MT evaluation metrics: 
\begin{itemize}
    \item (1) \emph{granularity} (thumbs-up vs. thumbs-down, or scoring on a 1-5 scale)
    \item (2) \emph{variance} (different users may rate the same output differently)
    \item (3) \emph{skew} (a user might be a harsher critic in general, and be prone to giving lower scores on average than other users)
\end{itemize}

In the experiments, we simulate use feedback by transforming raw SentenceBLEU scores into lower \emph{granularity} bins on a 1-5 rating scale (Step 4). In Section \ref{sec:bandit-losses}
 we further assess the effect of different simulated feedback to bandit learning.

\section{Experiments}

We aim to study the effectiveness of bandit algorithms on the task of MT system selection, across a variety of domains (and domain mixtures).  Here we introduce the MT systems, the datasets used to train them, and the bandit algorithms used to learn the system selection policy.

\subsection{Datasets}

Our experiment data consists of three different tasks, translating from German to English: 
\begin{enumerate}
\item The \textbf{General-Domain} task includes data from a range of domains, and is meant to be reflective of the kind of data used in public deployed systems. 
Specifically, we include OpenSubtitles2018 \citep{lison-tiedemann-2016-opensubtitles2016} and WMT 2017 \citep{bojar-etal-2017-findings}, which contains data from e.g. parliamentary proceedings (Europarl, UN), political/economic news, and web-crawled parallel corpus (Common Crawl). After filtering out long sentences ($>$80 tokens), we obtain a training set of 28 million sentence pairs. 
\item  The \textbf{TED} task focuses on translating captions from TED Talks, which contains specialized vocabulary in various professional fields (e.g. technology, entertainment, design) in the form of monologue speeches. We use the WIT3 data distribution \citep{cettoloEtAl:EAMT2012} with the  train/dev/test splits provided by \citet{duh18multitarget}.
\item The \textbf{WIPO} task focuses on patent translation, which contains even more specialized jargon, written in a formal style. We use the COPPA V2.0 distribution \citep{junczys-dowmunt2016coppa}. We held out 3000 random sentences each for dev and test, leaving 821 thousand sentences as training data.

\end{enumerate}

\subsection{MT Systems}

The training and development data described above are used to build machine translation systems. Bandit experiments are run on the test data, which has 1982, 3000, and 5504 sentences for the TED, WIPO, and General tasks respectively.  All data is tokenized by the Moses tokenizer~\citep{koehn2007moses}, then split into subwords by BPE~\citep{bpe} independently with 30k merge operations for the English and German sides. 

\paragraph{Neural machine translation} Models are built with Sockeye~\citep{hieber2017sockeye}, using common settings for LSTM seq2seq models~\citep{Bahdanau}: 2 layer encoder, 2 layer decoder, 512 hidden nodes, 512 word embedding sizes. 
\paragraph{Statistical machine translation} Models are built with Joshua~\citep{post-etal-2013-joshua}. 
This represents a strong phrase-based SMT model with GIZA++ alignments, 4-gram language model, and MIRA-based discriminative tuning. 

\paragraph{Systems} For each task, we train SMT and NMT models from scratch using only the training data in the respective domains, resulting in 6 models: \{nmt,smt\}-\{general,ted,wipo\}. 
Additionally we include two improved NMT models for WIPO and TED (nmt-cont-\{ted,wipo\}), which starts with nmt-general as initializaton and fine-tunes on WIPO or TED training data. This continued training process usually achieves strong translation performance in the target domain, but shows increased risk of catastrophic forgetting in the original general domain task \citep{luong-manning:iwslt15, thompson-etal-2019-overcoming}.  This brings the total number of systems (also the number of bandit arms, $K$) to 8.  The baseline performance of each system on the test sets in each domain is shown in Table~\ref{tab:arm-performance}.

\paragraph{Metrics} The bandit learning agents are updated on-the-fly on the aforementioned test sets, using perturbed/granularized sentence-level BLEU as feedback. However, for final evaluation we collect all the resulting translations and compute corpus-level BLEU \citep{bleu} (implemented via SacreBLEU \citep{sacrebleu}) and TER \citep{ter}. We experiment with different ways to mix the test sets to illustrate different scenarios for bandit learning. For error analysis, we computed regret as in Eq. \ref{eq:regret}.

\begin{table*}[t]
    \centering
    \begin{tabular}{r cc cc cc cc}
    \toprule
    & \multicolumn{2}{c}{\textbf{GENERAL}} & \multicolumn{2}{c}{\textbf{TED}} & \multicolumn{2}{c}{\textbf{WIPO}} & 
    \multicolumn{2}{c}{\textbf{ALL}} \\ 
       \cmidrule(lr){2-3}
       \cmidrule(lr){4-5}
       \cmidrule(lr){6-7} 
       \cmidrule(lr){8-9}
       & BLEU$\uparrow$ & TER$\downarrow$  & BLEU$\uparrow$ & TER$\downarrow$ & BLEU$\uparrow$ & TER$\downarrow$  & BLEU$\uparrow$ & TER$\downarrow$  \\
      \cmidrule(lr){2-2}
      \cmidrule(lr){3-3}
      \cmidrule(lr){4-4}
      \cmidrule(lr){5-5}
      \cmidrule(lr){6-6} 
      \cmidrule(lr){7-7}
      \cmidrule(lr){8-8} 
      \cmidrule(lr){9-9}
      nmt-general & \textbf{29.4} & \textbf{50.5} & 34.2 & 42.6 & 36.0 & 49.3 & 33.9 & 48.7 \\
smt-general & 23.9 & 54.9 & 30.7 & 45.8 & 26.7 & 56.9 & 26.5 & 53.8 \\
\cmidrule{2-9}
smt-ted & 16.5 & 62.3 & 28.7 & 47.9 & 12.0 & 69.5 & 15.6 & 61.6 \\
nmt-ted & 16.5 & 67.3 & 31.5 & 46.3 & 8.4 & 90.1 & 14.0 & 69.8 \\
nmt-cont-ted & 27.5 & 53.0 & \textbf{39.3} & \textbf{38.2} & 29.5 & 61.5 & 30.2 & 52.6 \\
\cmidrule{2-9}
smt-wipo & 9.9 & 79.3 & 9.7 & 77.5 & 51.2 & 36.0 & 35.2 & 66.6 \\
nmt-wipo & 6.6 & 101.2 & 7.7 & 92.1 & 61.9 & 25.4 & 39.0 & 77.8 \\
nmt-cont-wipo & 8.0 & 99.4 & 10.0 & 88.3 & \textbf{62.3} & \textbf{25.0} & 39.6 & 76.0 \\
    \bottomrule
    \end{tabular}
    \caption{Overview of translation  performance (DE $\rightarrow$ EN) for the eight systems which constitute the arms of the bandit.  Three architectures (nmt, nmt-cont, and smt) are trained and evaluated across three different domains.  Typically NMT with continued training (nmt-cont) is the highest performing system on in-domain data, but other systems offer more consistent performance.}
    \label{tab:arm-performance}
\end{table*}

\subsection{Bandit Methods}
\label{sec:bandit_methods}

We compare the three bandit methods:
\begin{itemize}
    \item \textbf{Epsilon-Greedy} The agent either exploits the arm with highest average reward with probability $1-\epsilon$, or chooses randomly (uniformly) with probability $\epsilon$~\citep{sutton-barto:1998}. 
    Intuitively, the agent maintains a running average of each arm's rewards; it greedily chooses the one with the highest reward, but occasionally tries a random arm in order to improve its estimate of the running average.
    
    \item \textbf{Upper Confidence Bound (UCB)}
    An $\epsilon$-greedy strategy is prone to obvious pitfalls.  For instance, if the optimal action performs poorly early on, it may take many iterations to correct the agent's behavior.  Alternatively, the agent can avoid becoming over-confident in its action reward estimates by establishing an upper confidence bound on each.  This encourages the model to explore actions which may have low empirical estimates of reward, if they have been tried infrequently.  The particular UCB algorithm we use in this work is UCB1~\citep{Auer:2002}.
    
    \item \textbf{Lin-UCB}
    Recall that in contextual bandits, the learner is presented with a feature vector, here derived from the source sentence.  The agent may use these feature vectors, along with the rewards of arms played in the past, to make a more informed choice of which arm to play at the current time step.  Over time, the aim of the learner is to collect enough information about how the context vectors and rewards relate to each other, so that it can predict the next best arm to play by looking at the feature vectors. 
    
    {\sc LinUCB} extends the principles of UCB to the contextual bandit scenario, using a linear model to predict the action~\citep{Li:2010:CAP:1772690.1772758}.  We explore various features for this system (results in Sec.~\ref{sec:features}) 
\end{itemize}

We also introduce three baseline systems: {\sc Random}, {\sc Oracle}, which always chooses the optimal translation system, and {\sc Best-Arm-Oracle}, which chooses single best system (the one which has the highest BLEU across the entire test set).

\section{Results}

\subsection{Overview: Comparison of Bandit Algorithms}

\begin{table*}[t]
\centering
\setlength{\tabcolsep}{4pt}
\begin{tabular}{@{\extracolsep{1pt}}r ccc ccc ccc ccc ccc}
\toprule   
 & \multicolumn{3}{c}{\textbf{GENERAL}}  & \multicolumn{3}{c}{\textbf{TED}} & \multicolumn{3}{c}{\textbf{WIPO}} &  \multicolumn{3}{c}{\textbf{AVG}} \\
 \cmidrule(lr){2-4} 
 \cmidrule(lr){5-7} 
 \cmidrule(lr){8-10}
 \cmidrule(lr){11-13}
 \cmidrule(lr){14-16}
& R$\downarrow$ & B$\uparrow$ & T$\downarrow$  & R$\downarrow$ & B$\uparrow$ & T$\downarrow$  & R$\downarrow$ & B$\uparrow$ & T$\downarrow$  & R$\downarrow$ & B$\uparrow$ & T$\downarrow$ \\
 \midrule
 random & 16.9 & 17.9 & 70.9 & 20.8 & 24.7 & 59.3 & 32.2 & 36.7 & 52.4 & 23.3 & 26.4 & 60.9\\
best-arm-oracle & 5.9 & 29.4 & 50.5 & 6.2 & 39.3 & 38.2 & 4.5 & 62.3 & 25.0 & 5.5 & 43.7 & 37.9\\
oracle & 2.2 & 31.6 & 49.9 & 1.8 & 42.1 & 37.2 & 2.3 & 61.7 & 24.1 & 2.1 & 45.1 & 37.0\\
\midrule
epsilon-greedy & 9.3 & 26.2 & 56.9 & 11.9 & 34.1 & 45.2 & 12.6 & 54.8 & 32.6 & 11.3 & 38.3 & 44.9\\
ucb & 9.9 & 25.4 & 56.7 & 13.2 & 32.7 & 46.8 & 9.3 & 58.0 & 29.8 & 10.8 & 38.7 & 44.4\\
linucb & \textbf{7.5} & \textbf{28.0} & \textbf{52.6} & \textbf{9.7} & \textbf{36.5} & \textbf{42.2} & \textbf{5.3} & \textbf{61.7} & \textbf{25.7} & \textbf{7.5} & \textbf{42.1} & \textbf{40.2}\\ 
\bottomrule
\end{tabular}
\caption{Results on the in-domain test sets.  Evaluation is measured in terms of average regret (R), BLEU (B), and TER (T). Lower scores are better for regret and TER; higher scores are better for BLEU.}
\setlength{\tabcolsep}{6pt}
\label{tab:bandit-in-domain}
\end{table*}

\begin{figure*}[t]
    \centering
    \includegraphics[scale=0.30]{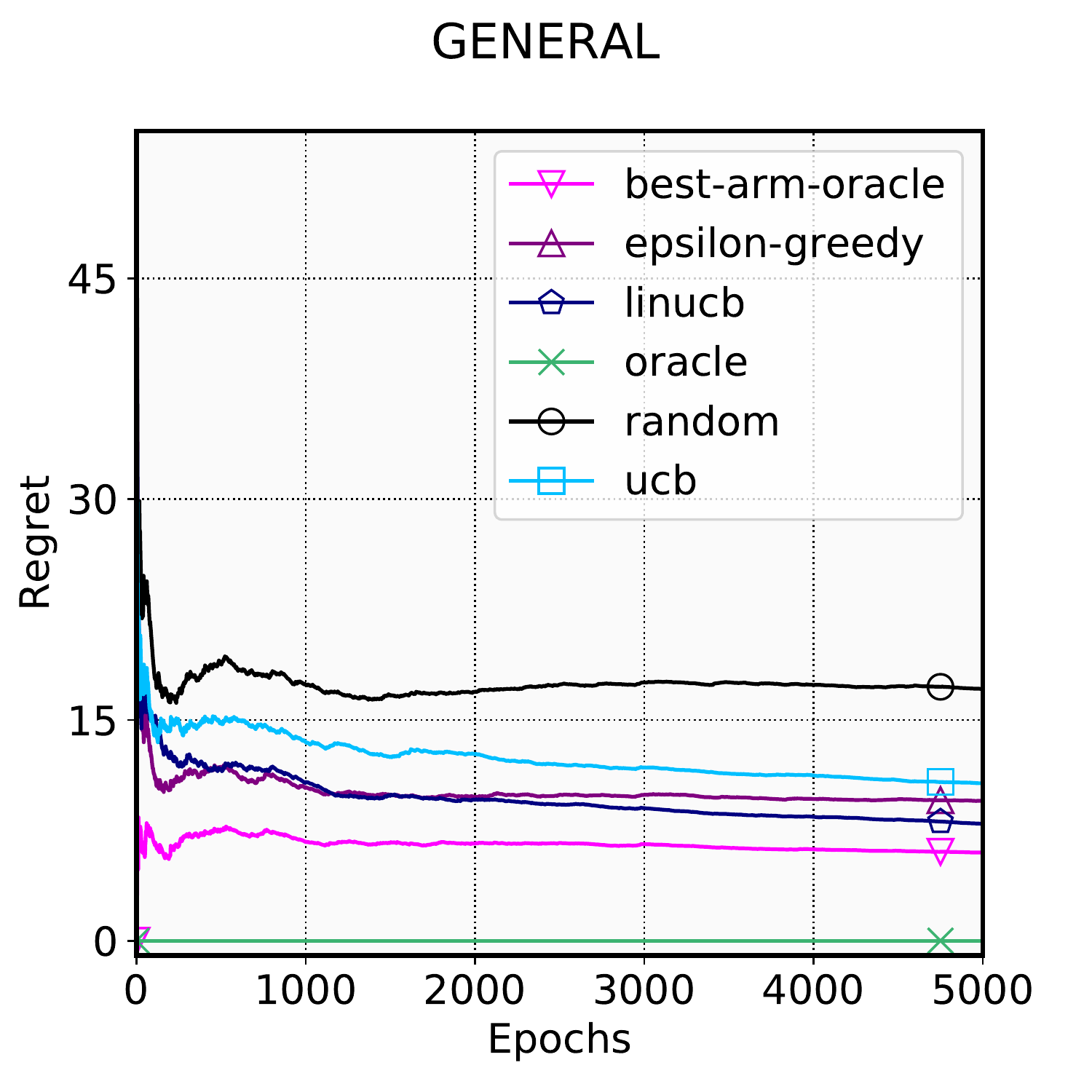}
    \hspace{-1em}
        \includegraphics[scale=0.30]{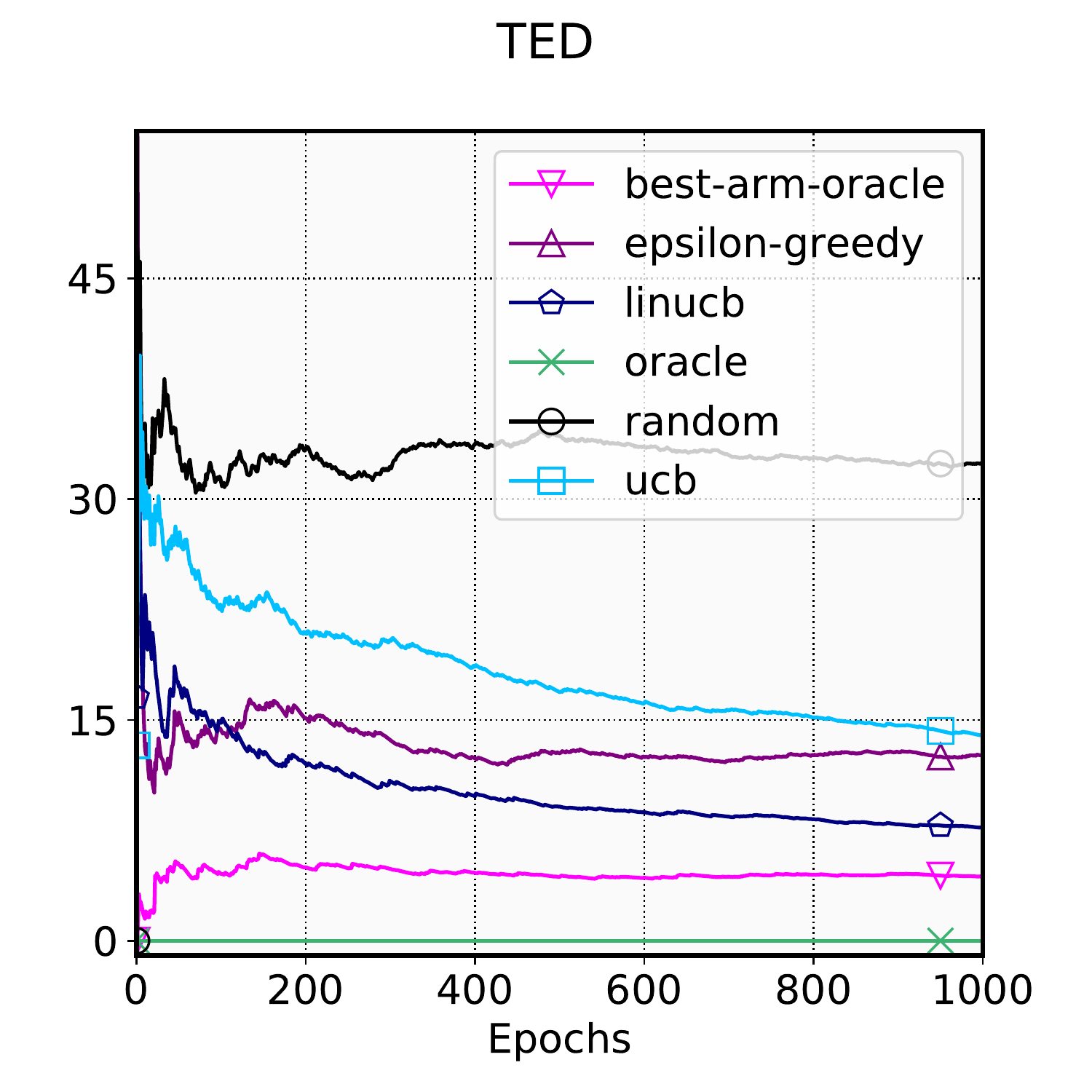}
            \hspace{-1em}
            \includegraphics[scale=0.30]{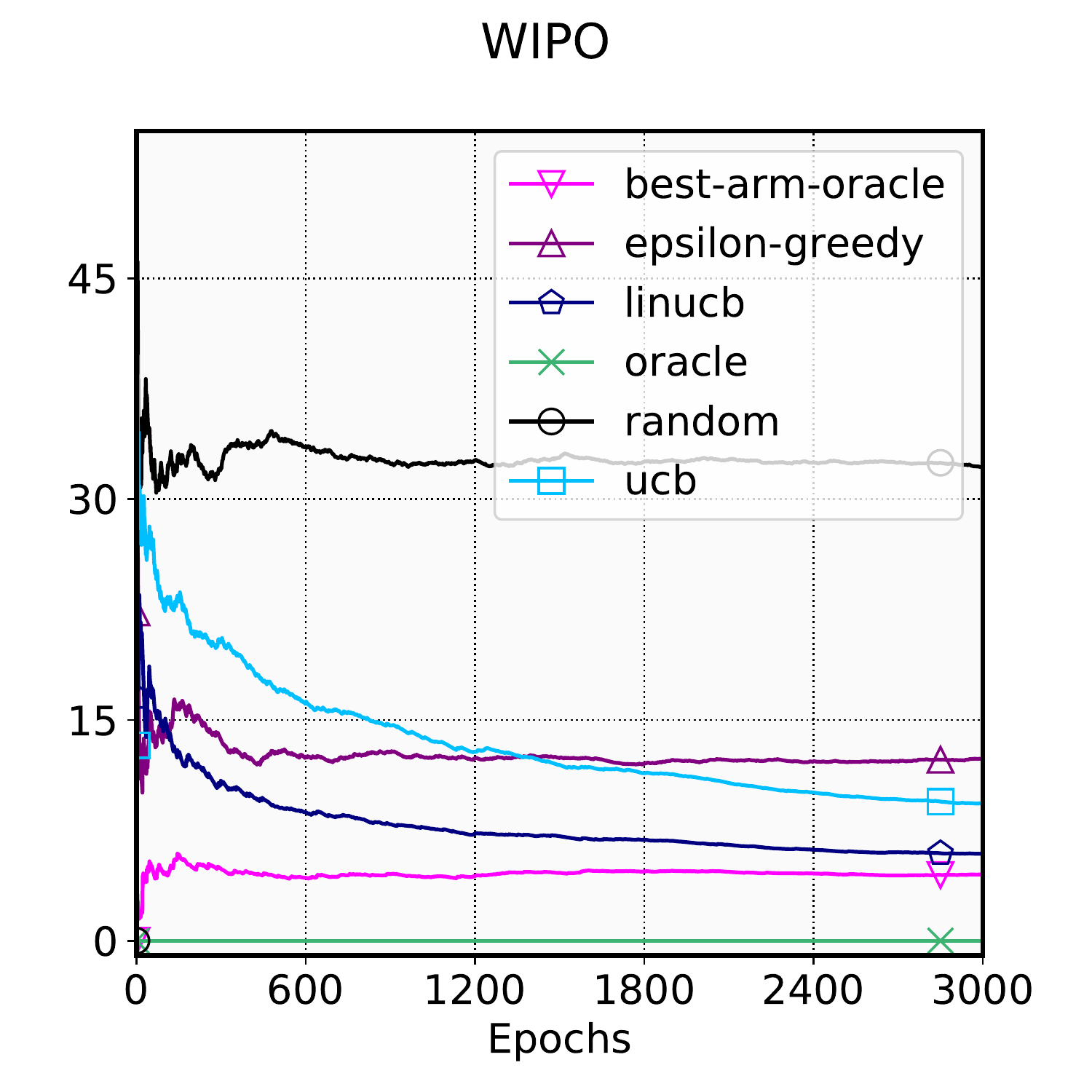}
    \caption{Comparison of cumulative regret in bandit algorithms across domains.  The relative ordering of the algorithms is consistent across all three domains, with epsilon-greedy adapting early, before eventually being surpassed by linucb, which converges closer to the best-arm oracle.  Note that cumulative regret lags behind changes in policy. Systems begin to find reasonable policies around epoch 50, marked by the sharp negative trajectory in cumulative regret.}
    \label{fig:regret-plots}
\end{figure*}

We begin by assessing the relative strengths of bandit algorithms on the translation system selection task, and examine how closely performance on the regret-based objective function corresponds to changes in translation evaluation measures like BLEU and TER.  In these experiments we run each algorithm on the full test data for each of the three domains.  As shown in Table \ref{tab:arm-performance}, the highest-performing system for each domain is always an in-domain variant of neural translation.  In this setting a good algorithm should quickly learn which arms are trained on in-domain data, and to exploit these as much as possible.  

Fig. \ref{fig:regret-plots} illustrates the overall performance of {\sc Epsilon-Greedy}, {\sc UCB}, and {\sc LinUCB}, with respect to two oracles and a random baseline.  We find that {\sc Epsilon-Greedy} is the fastest to converge. As the $\epsilon$ parameter balances exploration and exploitation, this behavior is somewhat within our control.  
Our results are obtained with $\epsilon = 0.3$. Empirically we observe that performance is quite robust to changes in $\epsilon$, and values in the range $0.2$ to $0.4$ result in negligible performance differences (comparable to changing the random seed). In comparison, {\sc UCB} performance follows a promising trajectory, but takes many more rounds to converge.  For the goal of tailoring translations based on user feedback, executing thousands of interactions is likely an exorbitant requirement.

Turning to {\sc LinUCB}, the contextual bandit algorithm, note the large gap between {\sc Oracle} and {\sc Best-Arm-Oracle} performance.  This is representative of the potential for additional performance gains when using a contextual method, even the data comes from a single domain.  Perhaps unsurprisingly then, we find that {\sc LinUCB} typically outperforms other bandit algorithms in later epochs, often closely approximating the performance of the {\sc Best-Arm-Oracle}.  However, {\sc LinUCB} also performs well in early epochs, and is often the best performing system after 100 epochs. Strong early performance means there may be little compromise to using a contextual approach like {\sc LinUCB}, even in single domain translation tasks.

It is also worth pointing out the small discrepancy between the cumulative regret and BLEU.  For instance, this is evident in the general domain results, where {\sc LinUCB} has lower cumulative regret, but a lower BLEU score.  While these two metrics are highly correlated in our experiments, there is a margin of disagreement and the ranking of systems can sometimes flip when comparing across these metrics.

\subsection{Adapting to new Domains}

A more realistic scenario may be a mixed-domain task, in which the user's translation needs are not fixed, but change over time.  We simulate this scenario by mixing data from each of the three domains in different ratios.  The main question we want to ask is: can the bandit algorithms outperform the {\sc Best-Arm-Oracle} system?  Doing so would be represent a clear advantage over deploying any single system.

\begin{table}[t]
    \centering
    \begin{tabular}{r ccc}
\toprule
& R$\downarrow$ & B$\uparrow$ & T$\downarrow$ \\
\midrule
random & 24.6 & 31.4 & 46 \\
best-arm-oracle & 17.9 & 34.2 & 42 \\
oracle & 0.0 & 57.9 & 32 \\
\hline
epsilon-greedy & 20.5 & 34.4 & 43 \\
ucb & 22.8 & 33.8 & 44 \\
linucb & \textbf{17.4} & \textbf{46.9} & \textbf{41} \\
\bottomrule
    \end{tabular}
    \caption{Performance on randomly shuffled data.  All bandit systems are able to adapt to new domains quickly enough to achieve performance comparable to choosing the single best system, but the contextual bandit significantly outperforms it.}
    \label{tab:my_label}
\end{table}

We present the performance of these systems in Table ~\ref{tab:my_label}.  The results show that as the data is increasingly mixed, the contextual bandit, {\sc LinUCB}, significantly outperforms any single system.  When data is completely shuffled, this amounts to a gain of more than $12$ BLEU over the single best system, a relative improvement of over $37\%$.  This is also true of simpler bandits, and {\sc epsilon-greedy} also outperforms the single-best system, but by a narrower margin.

\begin{figure*}[t]
\centering
    \includegraphics[scale=0.22]{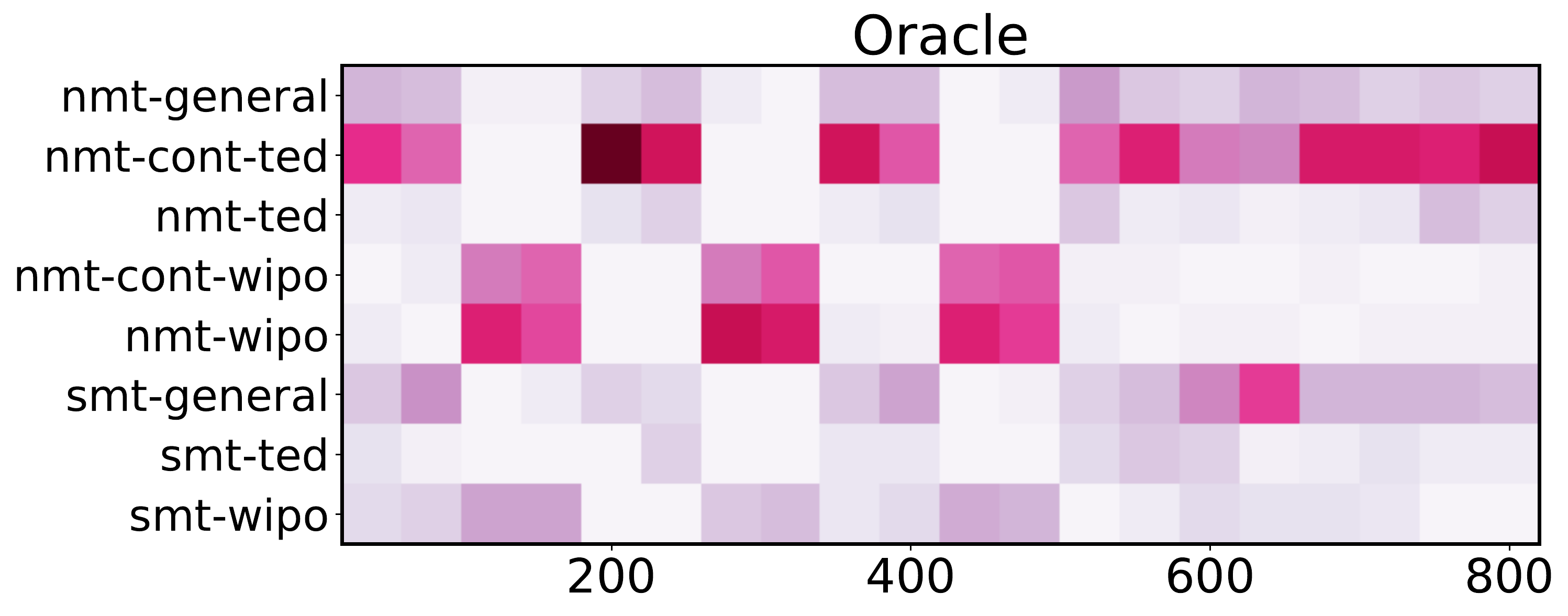}
    \includegraphics[scale=0.22]{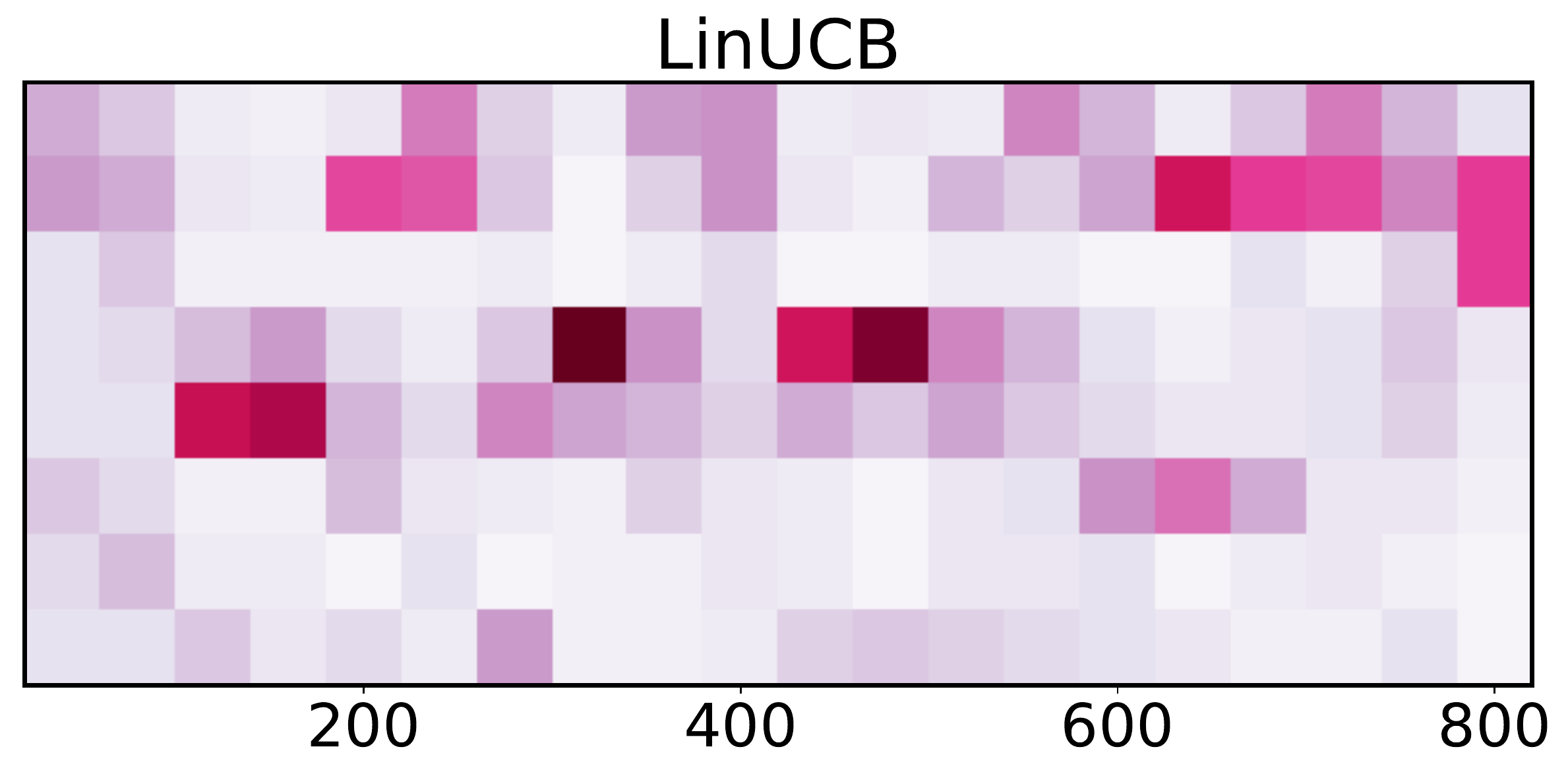}
    \includegraphics[scale=0.22]{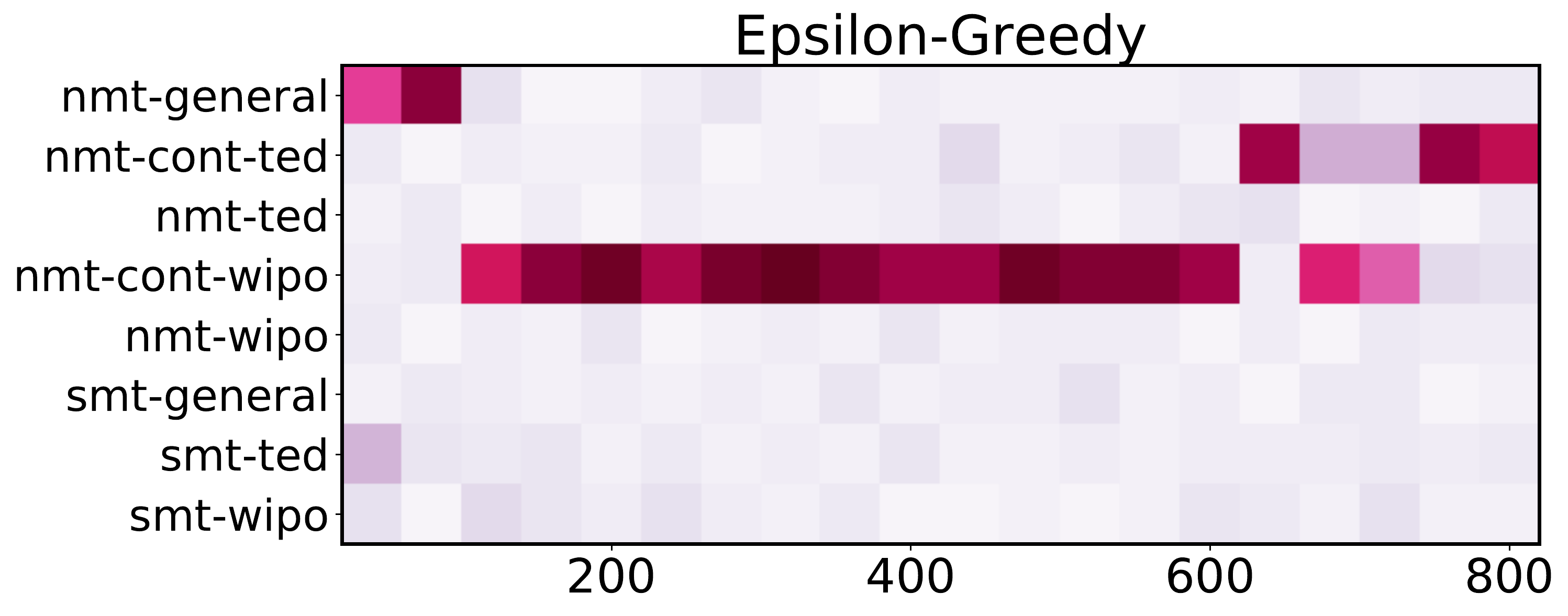}
    \includegraphics[scale=0.22]{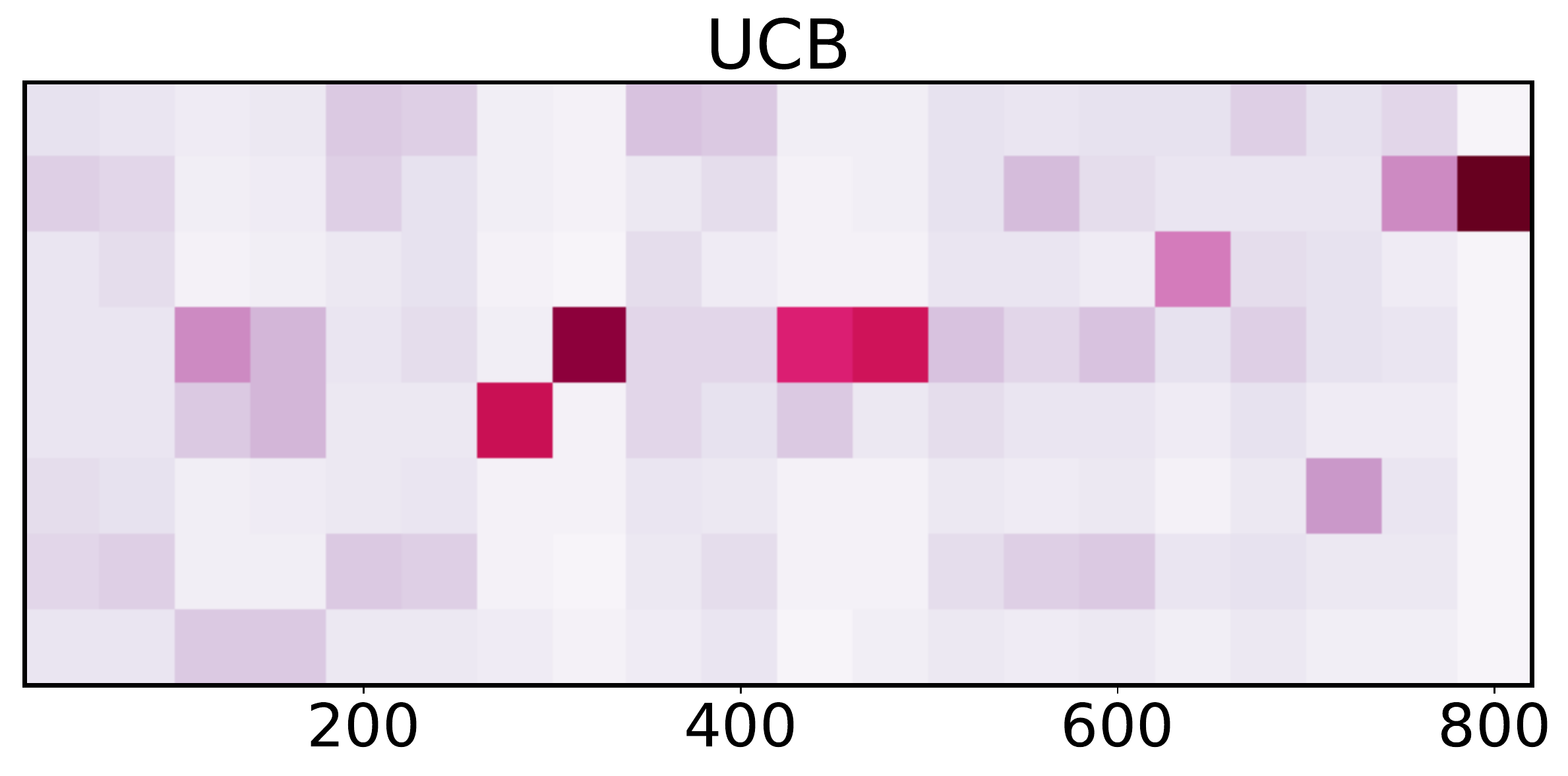}
\vspace{1em}
\vspace{1em}
    \includegraphics[scale=0.22]{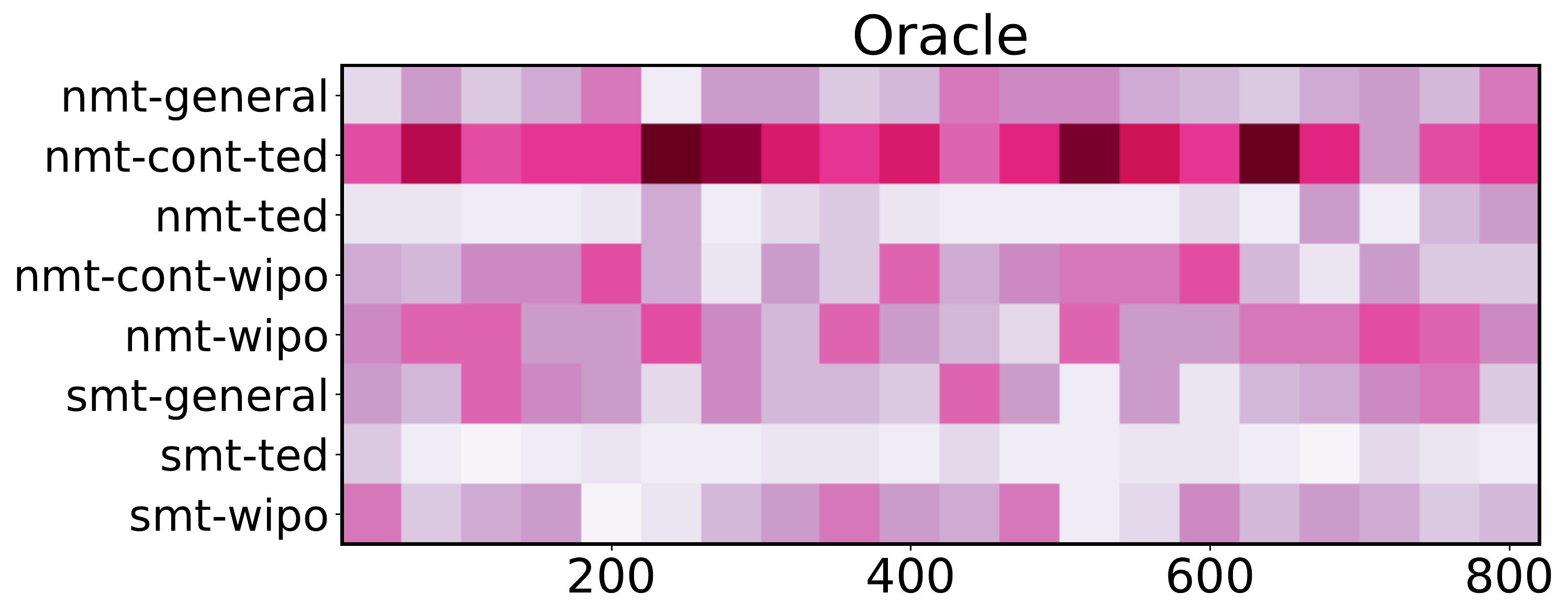}
    \includegraphics[scale=0.22]{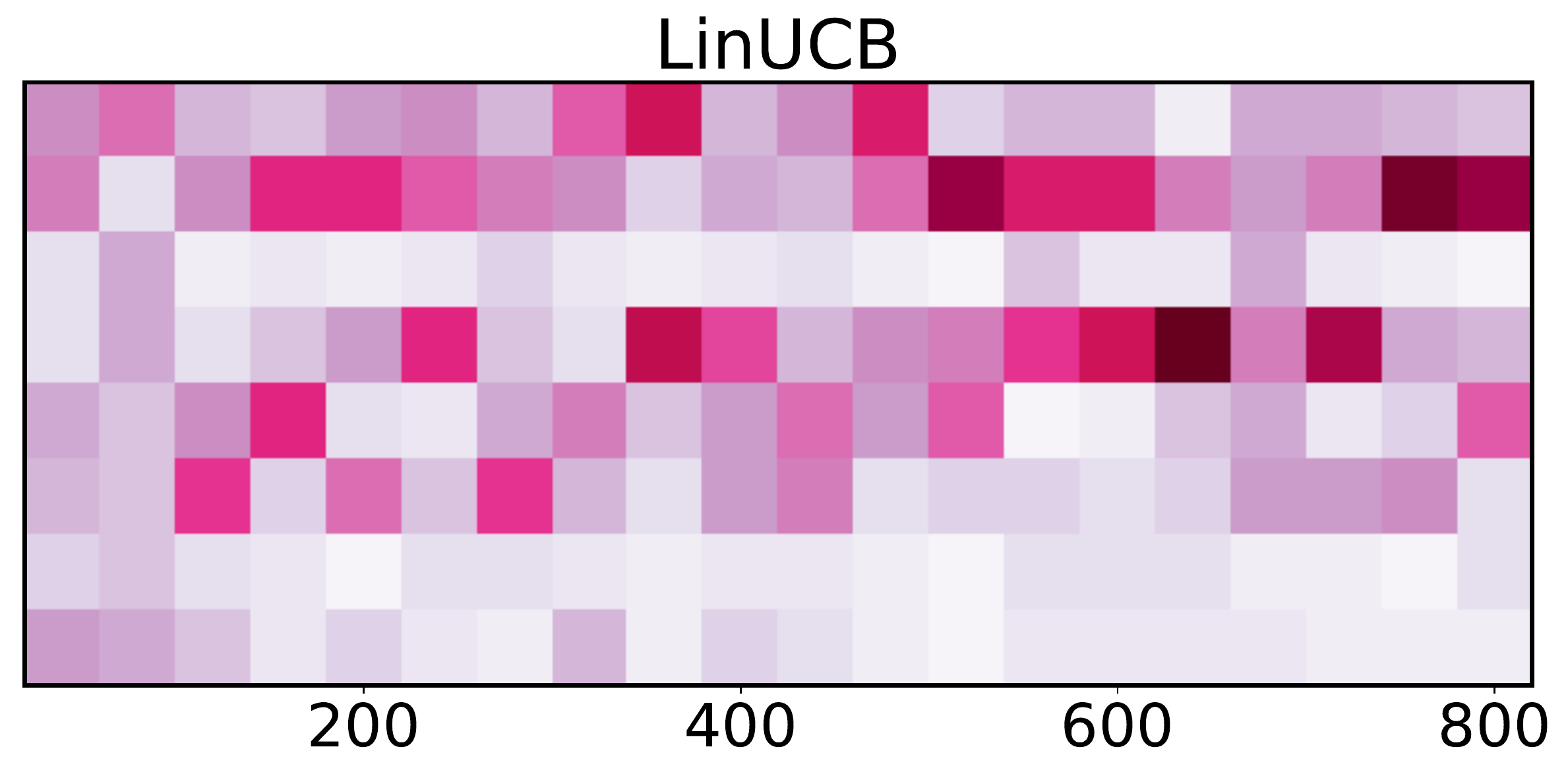}
    \includegraphics[scale=0.22]{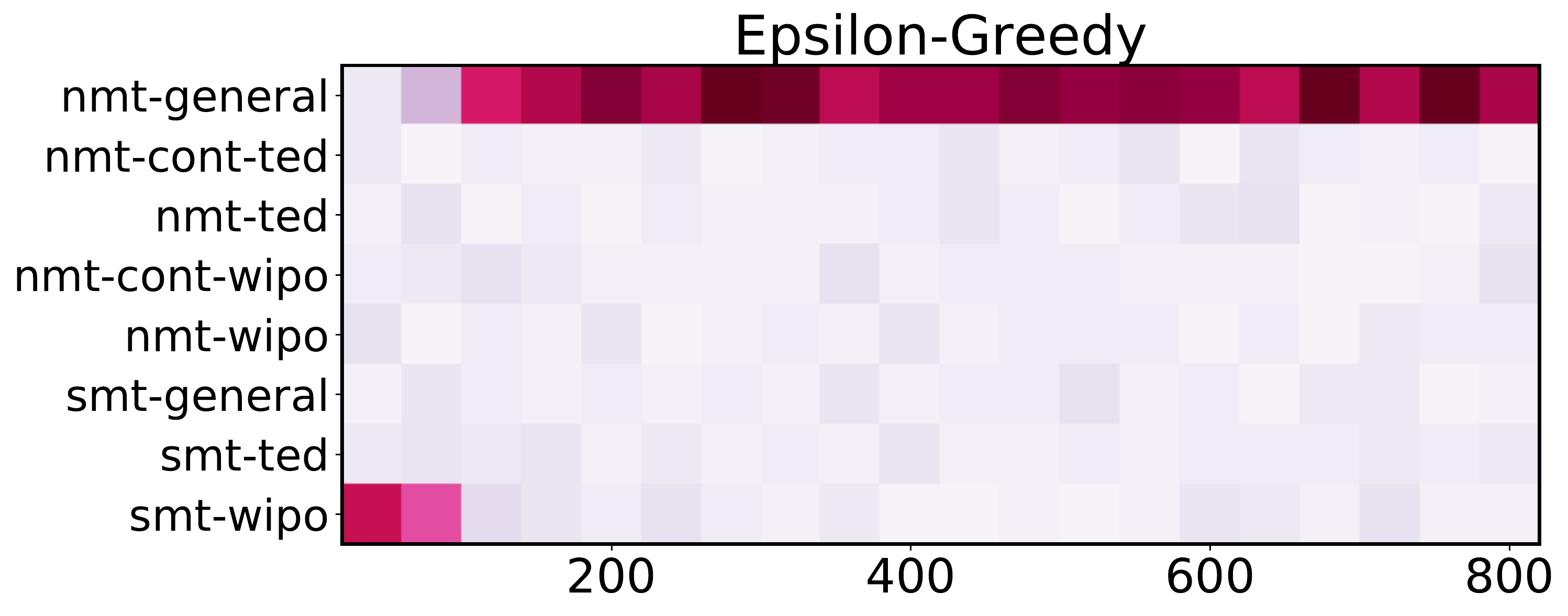}
    \includegraphics[scale=0.22]{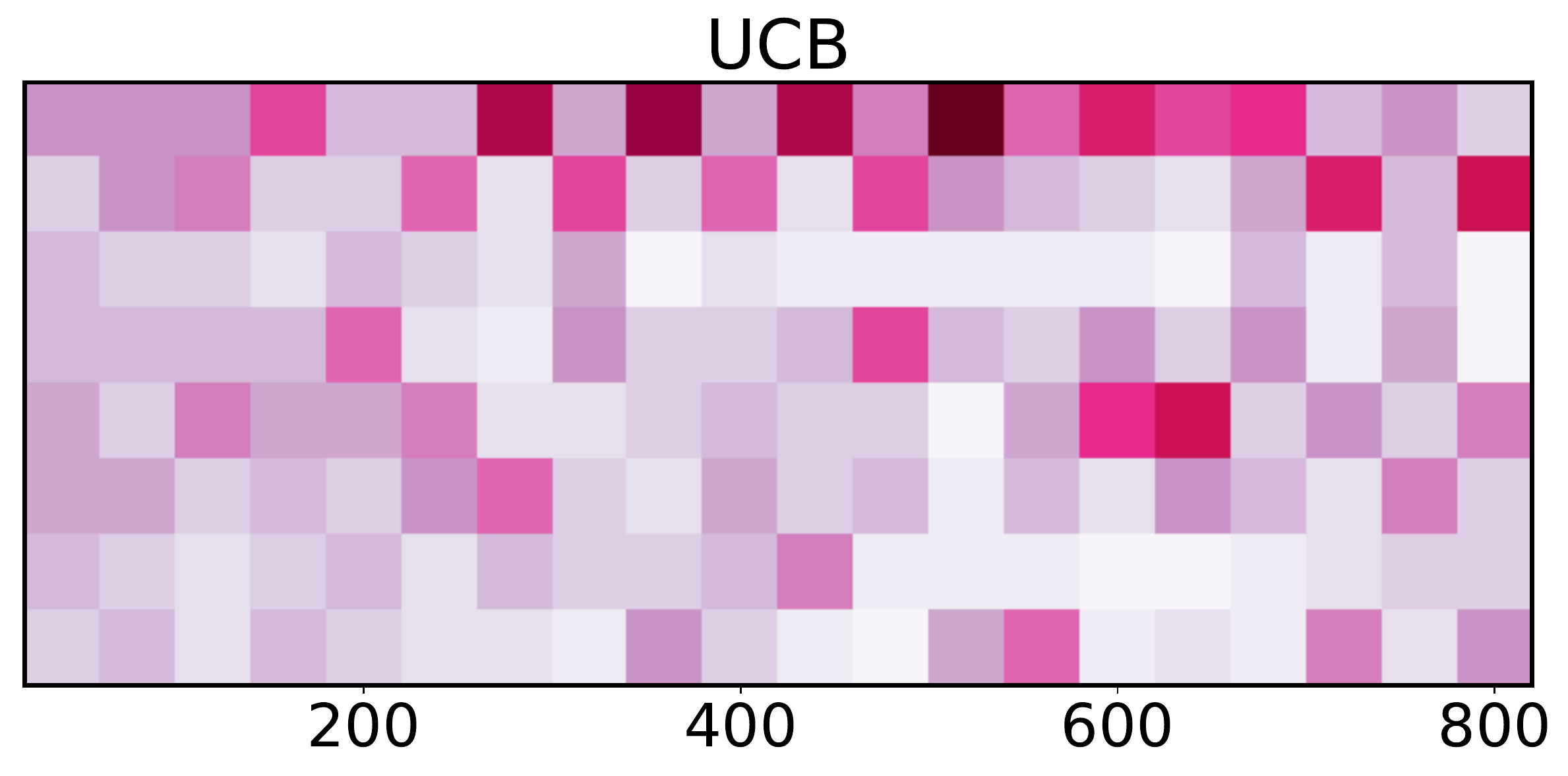}
    \caption{Decision heatmaps for shuffled data depicting the behavior of bandit algorithms across time.  Each column of the heatmap represents the distribution over the choices made by the agent during that interval of training (red squares correspond to actions taken more frequently).  In the top four plots we cycle the data domain every 100 examples.  In the bottom four plots, the data is drawn randomly from each of the three domains.  For clarity, we omit presenting heatmaps of the best-arm-oracle system, which is nmt-cont-ted in both scenarios.
    } 
\label{fig:mixed-domain-heatmaps}
\end{figure*}

Heatmaps of the algorithm decisions (Figure ~\ref{fig:mixed-domain-heatmaps}) provide some insight into the behavior of these systems.  In fully randomized sequences, we observe that after 100-200 iterations of learning, {\sc LinUCB} is able to closely mimic the behavior of the {\sc Oracle} system, ultimately converging to a similar distribution over decisions.\footnote{From the {\sc Oracle} heat map, we see that all 8 arms/systems are chosen at some point; this shows that even though we may have prior knowledge that one system is generally better than another, it is still useful to include all systems if selection is performed at the sentence level.}  {\sc epsilon-greedy} converges to predominantly choosing the second-best arm as a safe bet, while {\sc UCB}, which adjusts its policy more slowly, never exhibits clear decision trends.  

Even in less mixed scenarios (\ref{fig:mixed-domain-heatmaps}), top), results show that {\sc LinUCB} is an effective system.  As the domain changes between TED and WIPO, {\sc LinUCB} closely tracks the {\sc Oracle} decisions, even within the first 50 iterations, making the system a promising option for real-world deployment.

\subsection{Simulated Bandit Feedback}
\label{sec:bandit-losses}

In order to ascertain how sensitive bandit translation system selection is to the nature of simulated feedback, we explore different types of of constructing feedback.  
Recall the ways in which human feedback may differ from continuous metrics as identified in \citet{nguyen-daumeiii-boydgraber:2017:EMNLP2017} are granularity, variance, and skew. 

To simulate granular feedback, we bin SentenceBLEU scores into one of 5 equally-sized bins.  For variance in the feedback score, we sample the score from a Gaussian (with a variance shrinking parameter of $1.0$) that is updated after each evaluation.  For skew, we simulate a harsher critic which biases the output towards scores in a lower range (a skew of $0.25$).\footnote{Experiments in the previous section used granular feedback method.} For the sake of comparison we also include a scaling function, which simply scales the BLEU score into a suitable [0,1] loss range.  This serves as an ``oracle'' of what performance we might expect if we were somehow able to ask users to provide a fine-grained score like BLEU.

Table~\ref{tab:feedback-type} shows the effect of the four simulated feedback styles ({\sc Granular}, {\sc Variance}, {\sc Scale}, and {\sc Skew}) in a mixed data setting.  We find that the nature of feedback has little overall effect on system performance.  Altering the BLEU score to simulate {\sc Skew} was notably the most detrimental, but the remaining methods all performed similarly.  Surprisingly we observe no significant difference between {\sc Scale}, which is essentially the full continuous BLEU metric, and other distortions to the feedback score.

\begin{table}
    \centering
\begin{tabular}{c | c c c | c c c | c c c | c c c }
\toprule
& \multicolumn{3}{c|}{SCALE} & \multicolumn{3}{c|}{VARIANCE} & \multicolumn{3}{c|}{GRANULAR} & \multicolumn{3}{c}{SKEW} \\
 \midrule
 & R$\downarrow$ & B$\uparrow$ & T$\downarrow$ & R$\downarrow$ & B$\uparrow$ & T$\downarrow$  & R$\downarrow$ & B$\uparrow$ & T$\downarrow$ & R$\downarrow$ & B$\uparrow$ & T$\downarrow$ \\
epsilon-greedy & 19.3 & 35.0 & 42 & 21.6 & 37.5 & 44 & 19.0 & 37.3 & 42 & 19.7 & 33.1 & 43 \\
ucb & 13.3 & 50.1 & 38 & 13.3 & 50.7 & 38 & 13.1 & 50.2 & 38 & 12.3 & 51.0 & 38 \\
linucb & 14.1 & 48.9 & 39 & 13.5 & 48.9 & 39 & 13.6 & 48.5 & 39 & 15.6 & 45.2 & 40 \\
\bottomrule
\end{tabular}
\caption{Effects of simulated feedback method on  performance.}
    \label{tab:feedback-type}
\end{table}

\subsection{Features for Contextual Bandits}
\label{sec:features}

Contextual bandit methods make use of feature vectors when choosing an action.  In the context of translation, we can construct this vector from useful information in the source sentence.  We experiment with three types of features: 
\begin{itemize} 
\item OOV, whether the source sentence contains a high proportion of out-of-vocabulary words.
\item LEN, the length of the source sentence binned into five ranges of five (1-5, 6-10, ..., $>$25).\footnote{We use only the length feature for {\sc LinUCB} in earlier in-domain experiments~(Table~\ref{tab:bandit-in-domain}).}
\item BERT, features taken from a pre-trained language model~\citep{devlin-etal-2019-bert}.  Specifically, we ran the multilingual BERT base-size model out-of-the-box\footnote{\url{https://github.com/google-research/bert}} in inference mode and extract the final layer of the transformer encoder.  These embeddings are averaged across all tokens in the sentence. Since the evaluation datasets used in this study are small, on the order of thousands of examples, {\sc LinUCB} has difficulty learning quickly when using large feature vectors.  Therefore we take only the first 50 embedding dimensions as features. 
\end{itemize}

A constant bias feature is used to establish a baseline. Table~\ref{tab:bert-features} shows the results.  As evident by comparing against the {\sc Bias} feature results, all feature types provide useful information to the system selection task, though length and BERT features prove to be much more effective than vocabulary-based features.   

\begin{table}[h]
    \centering
        \begin{tabular}{l c c c}
        \toprule
         & R$\downarrow$ & B$\uparrow$ & T$\downarrow$ \\
        \midrule
        All & 13.6 & 47.7 & 40 \\
        \midrule
        \ \ OOV & 19.2 & 33.0 & 42 \\
        \ \ LEN & 16.8 & 45.9 & 41 \\
        \ \ BERT & \textbf{13.3} & \textbf{48.1} & \textbf{40} \\
        \ \ BIAS & 19.0 & 31.8 & 42 \\
        \bottomrule
        \end{tabular}
\caption{Ablation of contextual bandit features, on the randomly mixed-domain data.}
\label{tab:bert-features}
\end{table}



\section{Related Work}

\paragraph{Multi-domain machine translation}
Our problem setting is closely related to multi-domain machine translation, where the data comes as a stream of sentences from a mixture of domains unknown to the model~\citep{farajian-etal-2017-multi, Huck2015MixedDomainVM}. Typically using a single model, various extensions allow the system to cater more specifically to different types of source sentences.  Such extensions may include data concatenation, model stacking, data selection and multi-model ensemble~\citep{nmt-multi-domain:sajjad}.  One way is to pre-compute a domain label for each sentence using a dedicated classifier or model~\citep{kobus-etal-2017-domain, multi-domain:tars}.  

In neural models such adaptations can also be done in the learned representations.  \citet{britz-etal-2017-effective} use a discriminator network on top of the encoder to distinguish between domains and pretend a domain token to the target sequence.  \citet{gu-etal-2019-improving}
employ a shared encoder-decoder and also private models to capture both domain-invariant and domain-specific knowledge, which are then combined to generate the target sequence.  Such approaches can also be more nuanced, such as focusing on domain-specific words, and adjusting the training objective to emphasize their importance~\citep{zeng-etal-2018-multi}. 

A natural question is what advantage bandit system selection has over the alternative strategy of using of a domain classifier to determine which system to use.  One advantage of bandit selection is that it provides a unifying framework for online learning all aspects of problem.  If a domain classifier is useful, the classifier's predictions can be used as a feature in the bandit algorithm, and the extent to which that feature is useful (or useful together with other information) will also be learned.  Otherwise, the domain classifier features themselves can be incorporated into a contextual bandit policy where they can be access directly.

A second consideration is that the domains encountered in deployment may not be so well-defined and aligned to the translation systems as they are in our experimental setup.  Imagine a specialized domain that is not similar to the training domain of any system.  In these cases, other attributes of the model, such as its architecture or optimization strategy, may become more important.  This is a situation the bandit approach is well-suited for.

\paragraph{Bandits in NLP}
Due to the high cost of sourcing human annotations for NLP tasks, developing tractable training methods for learning from simple feedback has long been a desirable goal.  Learning NLP tasks (machine translation, sequence labeling, text classification) from bandit feedback has been studied previously~\citep{lawrenceSR17, sokolov-etal-2016-learning}, and has been extended to train typical NLP architectures, such as neural sequence-to-sequence models~\citep{Kreutzer:P17-1138}.  

Bandits have also been applied previously in MT, even as the topic of a dedicated shared task~\citep{sokolov:W17-4756}.
Within the context of bandit-driven MT, the focus has been on adapting an existing system, limited to simulated bandit feedback.  ~\citet{sokolov-etal-2016-learning} used actual losses (BLEU) and pairwise ranking. \citet{nguyen-daumeiii-boydgraber:2017:EMNLP2017}, also used bandit learning but to adapt a single neural MT system.  Our approach is significantly different, in that it focuses on the use of a bandit-trained policy for selection, rather than adaptation or in-domain training.

\section{Conclusion and Future Work}

As MT systems become widely deployed, catering translation output to user needs, whether through adaptation or system selection, will become an increasingly important problem.  In this work we showed that existing bandit algorithms are surprisingly effective at quickly adapting output to user needs, when the problem is phrased as one of selection.  Contextual bandit system selection methods frequently outperform the use of a single translation system, establishing this technique as promising solution for dynamically adapting to user translation needs.

While we did not explore more recent bandit methods, including Bayesian bandits, or hierarchical bandits, intuitively such methods would be a good fit for a large scale version of this study.  We assume arms are independent from one another, but they have dependencies both in terms of their architectures and in terms of their training data. One may also have prior knowledge of likely domains/arms, which can be incorporated in more advanced bandit methods. 

We note that there are many kinds of deployment scenarios, depending on factors such as (a) the number of domains, (b) how much each domain drifts, and (c) whether feedback comes from a single user or multiple users. 
We have presented a proof-of-concept to show the potential effectiveness of bandits, but the exact scenarios where these methods are most appropriate still require more exploration. 
For example, the non-contextual bandit setup described here is appropriate for a computer-assisted translation application where one professional human translator post-edits sentences from long document translations. Here the number of post-edits can serves as implicit feedback, and bandit methods like {\sc Epsilon-Greedy} are likely to converge in time (before the end of document) for the translator to start seeing benefits.\footnote{An interesting extension is to collect the resulting post-edited translations in order to adapt a personalized MT engine, which can then be one of many arms in the bandit, or to allow multiple human translators to share their translation memories.} 
On the other hand, the contextual bandit setup might be beneficial for an online MT service provider, where independent and unrelated translation requests may be interleaved. This can be viewed as another way to implement fast adaptation, but then one also needs to compare with the aforementioned multi-domain methods to decide the most suitable solution. 

\section{Acknowledgements}
We thank the anonymous reviewers for their valuable comments.  Artwork used in
Fig.\ref{fig:bandit} comes from TheNounProject\footnote{\url{https://thenounproject.com/}}, and the artists John Caserta, alifrio, and Herbert Spencer.

\bibliographystyle{apalike}
\bibliography{bandit_mt}

\end{document}